\documentclass[twocolumn,showpacs,preprintnumbers,amsmath,amssymb,superscriptaddress]{revtex4-2}
\usepackage{graphicx}
\usepackage{float}
\usepackage{placeins}
\usepackage{dcolumn}
\usepackage{bm}
\usepackage{hyperref}
\usepackage{color}
\newcommand{\Ltask}{\mathcal{L}_{\mathrm{task}}}
\newcommand{\bth}{\boldsymbol{\theta}}

\begin{document}

% Title of paper
\title{Soft Quantization: Model Compression Via Weight Coupling}

% You can use the thanks command to add acknowledgments or footnotes
% to the title if needed

% Author names and affiliations
% \author{Daniel T. Bernstein^{1,2}, Luca Di Carlo^{2,3}, and David Schwab^4}
\author{Daniel T. Bernstein}
\affiliation{Center for the Physics of Biological Function, Princeton, NJ, USA}
\affiliation{Lewis-Sigler Institute, Princeton, NJ, USA}
\author{Luca Di Carlo}
\affiliation{Center for the Physics of Biological Function, Princeton, NJ, USA}
\affiliation{Joseph Henry Laboratories of Physics, Princeton, NJ, USA}
\author{David Schwab}
\affiliation{Initiative for the Theoretical Sciences, CUNY Graduate Center, New York City, NY, USA}

% Add more authors as needed
% \author{Third Author}
% \affiliation{Third Institution}

% Collaboration name if applicable
% \collaboration{Collaboration Name}

% The abstract
\begin{abstract}
We show that introducing short-range attractive couplings between the weights of a neural network during training provides a novel avenue for model quantization. These couplings rapidly induce the discretization of a model's weight distribution, and they do so in a mixed-precision manner despite only relying on two additional hyperparameters. We demonstrate that, within an appropriate range of hyperparameters, our ``soft quantization'' scheme outperforms histogram-equalized post-training quantization on ResNet-20/CIFAR-10. Soft quantization provides both a new pipeline for the flexible compression of machine learning models and a new tool for investigating the trade-off between compression and generalization in high-dimensional loss landscapes.
\end{abstract}

% Keywords (optional)
\keywords{neural network compression, model quantization, loss landscape, mixed-precision}

% Main text begins here
\maketitle
\noindent\textit{Introduction.---}
\begin{figure*}
    \centering
    \includegraphics[width=1\linewidth]{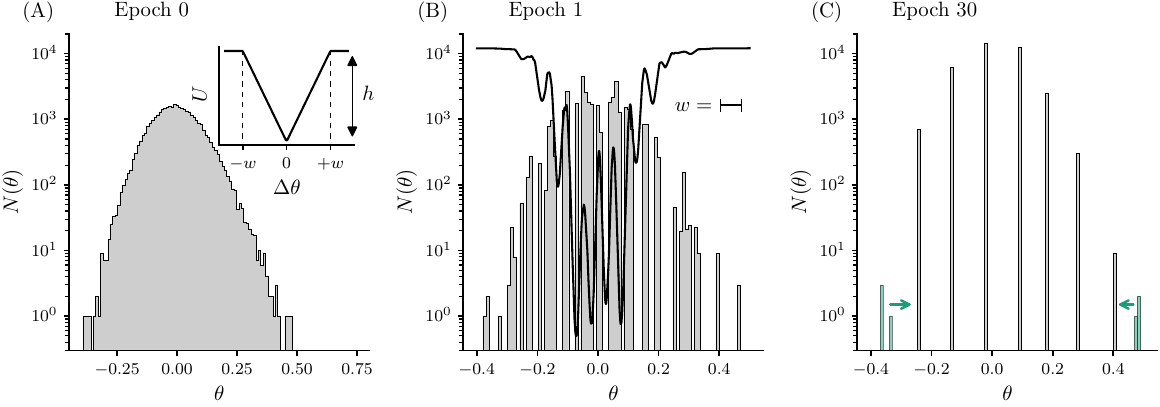}
    \caption{Weight distribution of a ResNet-20 layer during the soft quantization procedure of Table~\ref{tab:soft_quant_algorithm}. 
    (A) Distribution after pretraining on CIFAR-10; the inset shows a schematic of the potential between weight pairs.
    (B) Distribution after one epoch of soft quantization; the black curve shows the effective compression potential, Eq~\eqref{eq:effective_potential}, acting as a local L1 regularizer. 
    (C) Distribution at the end of the soft quantization; small clusters (in green) are reassigned to nearby larger ones, further reducing the bit-width of the layer.}
    \label{fig:schematic_fig}
\end{figure*}
Modern neural networks achieve striking performance, but at a memory and energy cost that grows rapidly with model size. Quantization can reduce this cost by representing parameters with a small set of discrete values, enabling lower-precision storage and arithmetic without changing network architecture \cite{nagel_white_2021,gholami_survey_2021,jacob_quantization_2017}. In standard post-training quantization (PTQ), the parameters of a trained model are mapped onto a reduced number of discrete values. Many models tolerate 8-bit PTQ with little loss in accuracy, while more aggressive compression typically requires careful bin placement or calibration \cite{nagel_white_2021,dettmers2022llmint8,dettmers2022optimizers}. Histogram-equalized quantization (HEQ) is a representative PTQ approach that chooses bins in a quantile-aware manner \cite{zhou_balanced_2017,nguyen_histogram-equalized_2022}. Quantization-aware training (QAT) can potentially further reduce quantization error by incorporating low-precision effects during optimization. Numerous QAT algorithms have been proposed in the literature~\cite{gholami_survey_2021,nagel_white_2021,yin2018understanding, BinaryRelax, jin_parq_2025}; this approach generally relies on approximations such as the straight-through estimator because the quantization map is non-differentiable \cite{yin2018understanding,BinaryRelax,jin_parq_2025}. 
Beyond algorithmic performance, quantization offers a window into loss-landscape geometry: how close are highly compressed solutions to task-optimal minima, and along which directions can weights move while incurring only a small increase in task loss? One way to probe these questions is to augment the task loss with a soft constraint that favors compressible configurations and then study the resulting trade-off between performance and compression as the constraint is tuned \cite{carlo2025quantization,ullrich2017soft,nowlan1992simplifying}.

In this Letter we explore \emph{soft quantization}, a fine-tuning procedure in which weights within each layer experience a short-range attractive interaction during training. The interaction causes nearby weights to aggregate into clusters, producing a discrete distribution and, generically, layer-dependent effective bit-widths controlled by two global hyperparameters. We demonstrate the approach on ResNet-20 trained on CIFAR-10 \cite{he2016deep,krizhevsky2009learning}, comparing against HEQ applied to the same pretrained model.

\noindent\textit{Framework.---}
For a task loss $\Ltask(\bth;X,Y)$ defined over data $\{X,Y\}$, we minimize
\begin{equation}
\mathcal{L}(\bth)=\Ltask(\bth;X,Y)+\mathcal{L}_C(\bth),
\label{eq:wwcompression}
\end{equation}
with a layerwise coupling term
\begin{equation}
\mathcal{L}_C(\bth)=\sum_{l} h_l \sum_{i \neq j} U_{w_l}\!\left(\theta_i^{(l)} - \theta_j^{(l)}\right),
\label{eq:compression_term}
\end{equation}
where $\bth^{(l)}=\{\theta_i^{(l)}\}_{i=1}^{N_l}$ are the weights in layer $l$, and $\{h_l,w_l\}$ set the interaction strength and range. We choose a short-range triangular well
\begin{equation}
U_{w}(x) = (|x|-w)\Theta(w^2-x^2),
\end{equation}
which is attractive for $|x|<w$ and vanishes otherwise (additive constants are irrelevant). We choose this form because it generates a constant attractive force for $|x|<w$ that rapidly attracts nearby weights while keeping the interaction strictly local \cite{tibshirani_regression_1996}. The range $w_l$ sets the scale over which nearby weights are encouraged to bind. In the large-$h_l$ limit, configurations in which two weights remain within a distance $w_l$ yet take distinct values become energetically unfavorable; during optimization such nearby weights rapidly coalesce, leaving separated clusters with typical spacings $\gtrsim w_l$.

A network with $L$ layers would in principle require $2L$ hyperparameters. To reduce this to two global values, we tie the interaction scales to layer statistics measured at the pretrained state. We set $w_l = w\,\sigma_l$, where $\sigma_l$ is the standard deviation of weights in layer $l$, and choose $h_l=h\,N_l^{-\alpha}$. Empirically, at pretrained initialization the unweighted interaction energy scales with layer size as $\sum_{i\neq j}U_{w_l}(\theta_i^{(l)}-\theta_j^{(l)})\propto N_l^{1+\alpha}$ with $\alpha\simeq0.66$ (see End Matter), motivating $h_l=h\,N_l^{-0.66}$. We treat this exponent as a practical scaling choice, and leave more rigorous scaling arguments to future work.

Direct evaluation of Eq.~(\ref{eq:compression_term}) requires summing over all pairs of weights in a layer and hence scales as $\mathcal{O}(N_l^2)$, which for layers with millions of parameters is prohibitive. Instead we work with the empirical weight density
\begin{equation}
\rho_l(\theta)=\frac{1}{N_l}\sum_{i=1}^{N_l}\delta\!\left(\theta-\theta_i^{(l)}\right),
\end{equation}
and define an effective potential
\begin{equation}
V_{w_l}(\theta)=N_l\int d\theta'\,\rho_l(\theta')\,U_{w_l}(\theta-\theta'),
\label{eq:effective_potential}
\end{equation}
by convolving the empirical weight distribution with the triangular well, so that
\begin{equation}
\mathcal{L}_C(\bth)\approx \sum_l h_l\sum_i V_{w_l}\!\left(\theta_i^{(l)}\right).
\end{equation}
The contribution from the compression-force to the gradient is thus
\begin{equation}
\frac{\partial \mathcal{L}_C}{\partial \theta_i^{(l)}}\approx h_l\left.\frac{dV_{w_l}(\theta)}{d\theta}\right|_{\theta=\theta_i^{(l)}}.
\end{equation}
Although $U_w$ is formally non-smooth, the binned estimate of $V_{w_l}$ is well-behaved in practice. We further regularize the computation by assigning each weight to the midpoint of its histogram bin when evaluating $V_{w_l}$ and its derivative, preventing numerical instabilities when many weights become extremely close.

In practice we approximate $\rho_l$ by a histogram with $N_b = 2^{14}$ bins spanning the support of $\bth^{(l)}$. The computational cost is then $\mathcal{O}(N_l)$ to construct the histogram and $\mathcal{O}(N_b \log N_b)$ to compute the convolution, yielding a substantial reduction in complexity from $\mathcal{O}(N_l ^2)$ to $\mathcal{O}(N_l + N_b \log N_b)$. In Figure \ref{fig:schematic_fig}(B) we show the effective potential (black curve) computed following this procedure. To add stochasticity in the compression term early in training, we construct the histogram from a random subsample of weights and increase the sampling fraction over epochs until the full layer is used by the final epochs. In our implementation, we calculate the compression-force directly rather than backpropagate through $V_{w_l}(\theta)$.

We apply this interaction during a fine-tuning of a pretrained model. We emphasize that we do not vary the data here, and this fine-tuning phase is strictly for our compression. Practically, this targets the common use case where one compresses an existing pre-trained network. Conceptually, it probes whether quantized configurations lie near task-optimal regions and are accessible by local deformations of the loss landscape. Figure~\ref{fig:schematic_fig} illustrates schematically the weight distribution throughout the entire soft quantization pipeline, as described in Table~\ref{tab:soft_quant_algorithm}.
To quantify compression we compute an \emph{effective bit-width} from the number of distinct weight clusters in each layer. After fine-tuning, we identify clusters in layer $l$ by binning the weight distribution (7-bit binning) and assigning weights to the mean value of weights in their bin. This yields $K_l$ distinct values (clusters) in layer $l$, and we define the layer bit-width as $b_l\equiv \log_2 K_l$. We report the parameter-weighted mean bit-width
\begin{equation}
\bar{b} \equiv \frac{\sum_l N_l\, b_l}{\sum_l N_l}.
\end{equation}
Because $\bar{b}$ depends on $K_l$ rather than directly on $N_l$, a small number of tiny clusters can inflate $\bar{b}$ while corresponding to a tiny fraction of parameters. We therefore optionally apply a simple refinement step in which clusters containing $\leq n_{\min}$ weights (we use $n_{\min}=10$) are reassigned to the nearest larger cluster after training; unless stated otherwise, reported bit-widths include this refinement.

\begin{table}[t]
\caption{Soft quantization procedure.}
\label{tab:soft_quant_algorithm}
\begin{ruledtabular}
\begin{tabular}{p{0.95\columnwidth}}
\textbf{Input:} pretrained weights $\{\bth^{(l)}\}$; global hyperparameters $(h,w)$; bins $N_b$. \\[2pt]
1. For each layer $l$, measure $N_l$ and $\sigma_l$ on the pretrained weights; set $w_l=w\,\sigma_l$, $h_l=h\,N_l^{-0.66}$. \\[2pt]
2. For each training step: \\
\quad (a) Build a histogram estimate of $\rho_l(\theta)$ using $N_b$ bins. \\
\quad (b) Compute $V_{w_l}(\theta)=N_l(\rho_l * U_{w_l})(\theta)$. Differentiate the binned $V_{w_l}$ to obtain $dV_{w_l}/d\theta$. \\
\quad (c) For each weight $\theta_i^{(l)}$, evaluate the compression-force term $h_l\,dV_{w_l}/d\theta$ at its bin midpoint and add to the SGD update. \\[2pt]
3. After fine-tuning, identify weight clusters (binning) and compute effective bit-widths. \\[2pt]
4. \textbf{Optional refinement:} merge very small clusters ($\leq n_{\min}$ weights) into the nearest larger cluster. \\[2pt]
\textbf{Output:} a compressed model with emergent clustered weights and layer-dependent effective bit-widths. \\
\end{tabular}
\end{ruledtabular}
\end{table}

\begin{figure*}[!t]
    \centering
    \includegraphics[width=\linewidth]{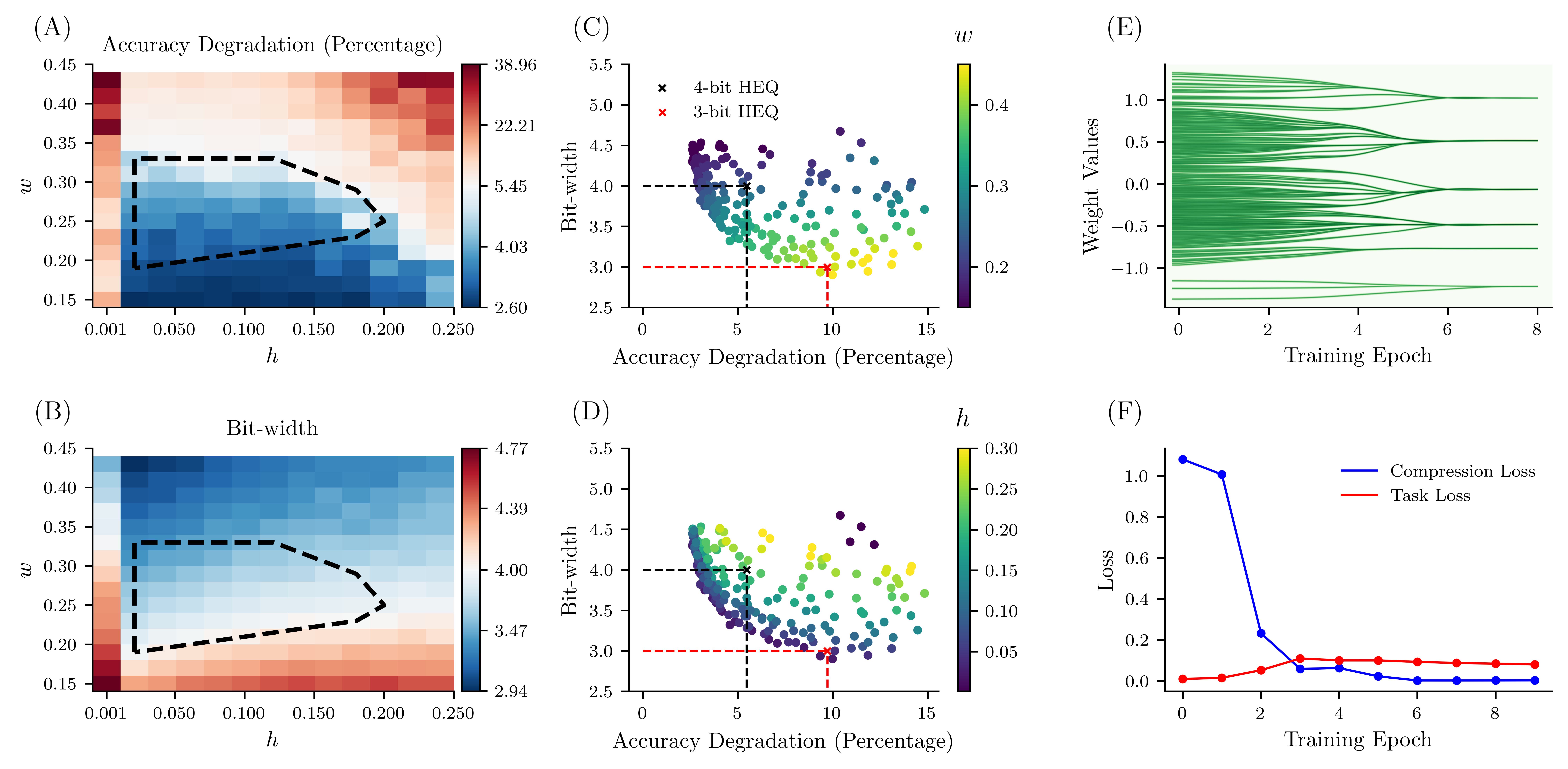}
    \caption{ 
    Performance of soft quantization compared with HEQ.
(A,B) Test accuracy degradation (A) and resulting bit-width (B) of softly quantized models as functions of the potential strength $h$ and range $w$. Colormap midpoints correspond to a 4-bit HEQ-compressed model. The black envelope outlines $(h,w)$ pairs that outperform 4-bit HEQ, achieving higher accuracy at lower bit-width. Each pixel represents an average over 15 independent simulations; standard deviations are small (see Fig.~\ref{fig:coeff_var} in the End Matter).
(C,D) Scatter plots of bit-width versus accuracy degradation for a range of models, compared with 4-bit and 3-bit HEQ benchmarks. Smaller $h$ values approach the Pareto-optimal frontier (D), while smaller $w$ values yield higher bit-width and lower degradation (C). Poor performance at $h = 0.001$ coincides with failure of the clustering refinement step, which reassigns small ($\leq 10$ weights) clusters when the coupling is too weak. Consistent with this, Figure ~\ref{fig:clamped_sq} in the End Matter shows that soft quantization effects are minimal for small $h$.
(E) Dynamics of final-layer weights during soft quantization. (F) Loss curves during soft quantization: compression loss (blue) rapidly decreases and saturates, while task loss (red) initially increases before gradually stabilizing.}
    % \caption{(A) Test accuracy degradation ($\Delta A = A_{\mathrm{pre}}-A_{\mathrm{comp}}$, in percentage points) and (B) mean bit-width of softly quantized models across values of $h$ and $w$. Colormap midpoints correspond to the performance of a 4-bit HEQ-compressed model. Each heatmap value is averaged over 15 simulations with the same hyperparameter settings. The black envelope outlines the hyperparameter settings which outperform 4-bit HEQ both in accuracy degradation and compression. The reason that we observe poor performance for $h = 0.001$ stems from the refinement process of reassigning small ($\leq 10$ weights) clusters, which leads to catastrophic degradation when the coupling is too weak and clustering fails during training. Figure \ref{fig:clamped_sq} in the End Matter corroborates this claim, showing that, for small $h$, the effects of soft quantization are minor. (C)-(D) Scatterplot of bit-width versus accuracy degradation for a range of models, compared against 4-bit and 3-bit HEQ benchmarks. (E) Dynamics of final-layer weights during soft quantization. Clusters form sharply at around the same period of time. (F) Loss curves during soft quantization. Compression loss (blue) drops sharply and levels out early in training, while task loss (red) rises before gradually tapering off.}
    \label{fig:Main_Scatterplot}
\end{figure*}

\noindent\textit{Results and Analysis.---}
We implemented our soft quantization procedure on ResNet-20 trained on CIFAR-10 \cite{he2016deep,krizhevsky2009learning}, using SGD with learning rate $0.001$ and Nesterov momentum $0.9$ for a total of 30 epochs. Our training loss is the standard cross-entropy loss for image classification. \cite{mao2023crossentropylossfunctionstheoretical}. We compare against HEQ applied to the same pretrained model. We emphasize that HEQ does not represent a state-of-the-art benchmark, but rather a reasonable and flexible ``first approach" to quantization. We report \emph{test accuracy degradation} as $\Delta A \equiv A_{\mathrm{pre}}-A_{\mathrm{comp}}$ (in percentage points), where $A_{\mathrm{pre}}$ is the test accuracy of the pretrained full-precision model and $A_{\mathrm{comp}}$ is the accuracy after compression.

We find that soft quantization outperforms HEQ across a range of $h$ and $w$ values, as shown in Figure \ref{fig:Main_Scatterplot}(A)--(B). For hyperparameter settings which outperform 4-bit HEQ, we observe low coefficients of variation in both accuracy degradation and bit-width (see End Matter), indicating that the compression outcome is consistent despite stochasticity in training dynamics.

Figure \ref{fig:Main_Scatterplot}(C)--(D) offer an interpretation for how varying $h$ and $w$ affects the trade-off between compression and performance. Lower values of $h$ place us closer to the Pareto frontier of compression and accuracy (although exceedingly small values of $h$ are insufficient for clustering, leading to poor performance after model refinement); $w$ controls whether we favor compression or performance, with small values of $w$ favoring lower accuracy degradation and higher values of $w$ favoring stronger compression.

As shown in Figure \ref{fig:Main_Scatterplot}(E), clusters form within the first few epochs, and the process of clusters collapsing to a single value happens at approximately the same time for the majority of clusters. The soft quantization procedure is characterized by a rapid drop in the compression potential concurrent with a rise in the training loss, followed by a steady and gradual drop in the training loss as the compression loss remains relatively flat (see Figure \ref{fig:Main_Scatterplot}(F)). This behavior suggests that the compression term allows the network to pass over small barriers in the loss landscape near pretraining configurations, finding configurations of low task loss which are also quantized.

\begin{figure}
    \centering
    \includegraphics[width=\linewidth]{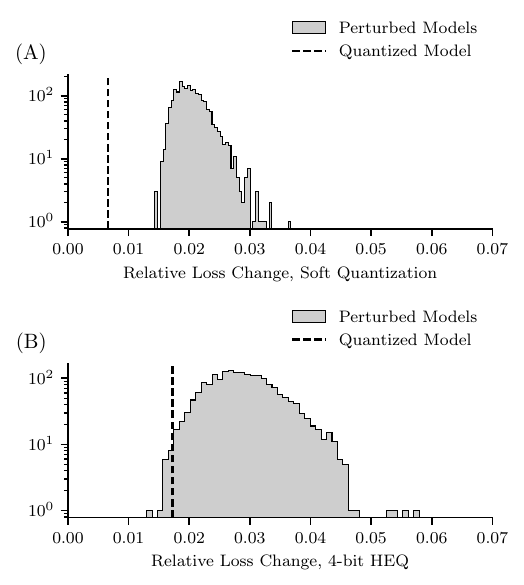}
    \caption{Relative task loss change $\frac{\Delta\Ltask}{\Delta\bth}$, for compressed and randomly perturbed models. 
    Here $\Delta \theta$ denotes the metric distance from the pretrained model and $\Delta \Ltask$ the corresponding change in task loss. Randomly perturbed models are obtained by adding layer-wise Gaussian noise, matching the parameter-space distance of the quantized models from the pretrained one. 
    (A) Soft quantization produces a significantly smaller $\frac{\Delta\Ltask}{\Delta\bth}$ than random perturbations at comparable distances. 
    (B) In contrast 4-bit HEQ  gives values of $\frac{\Delta\Ltask}{\Delta\bth}$ within the tail of the distribution. The softly quantized model shows a bit-width of $\approx3.79$ while being $50\%$ farther from the pretrained model than the 4-bit HEQ model. 
    %The softly quantized model shows an average bit-width of $\approx3.79$, and the distance it travels in the loss landscape is $48\%$ farther than the 4-bit HEQ model. 
    % between quantization and random perturbations for (A) soft quantization (before model refinement) and (B) 4-bit HEQ. Perturbations consist of displacing weights by white noise comparable to the weight displacement incurred by the corresponding quantization setting. We run 2000 perturbations for each setting. Soft quantization incurs a much smaller increase in $\frac{\Delta\Ltask}{\Delta\bth}$ than expected by a random perturbation, whereas 4-bit HEQ incurs a value of $\frac{\Delta\Ltask}{\Delta\bth}$ which is within the tail of the perturbation distribution. The softly quantized model shown has an average bit-width of $\approx3.79$, and the distance it travels in the loss landscape is $48\%$ farther than the 4-bit HEQ model.
    }
    \label{fig:perturbations}
\end{figure}
We analyze the efficiency with which soft quantization (before refinement) traverses the loss landscape by comparing $\frac{\Delta\Ltask}{\Delta\bth}$ in a softly quantized model to models which have been randomly perturbed after pretraining. To generate randomly perturbed models comparable to the softly quantized model, we add independent Gaussian noise of standard deviation $\eta_l = \frac{\Delta \bth_l}{\sqrt{N_l}}$ to each weight in layer $l$, where $\Delta \bth_l$ is the L2 distance between the pretrained and quantized weights in layer $l$. On average, the randomly perturbed weights will be displaced per layer by a distance $\Delta \bth_l$. As a benchmark, we carry out the same procedure for a model quantized with 4-bit HEQ.

As illustrated by Figure \ref{fig:perturbations}, $\frac{\Delta\Ltask}{\Delta\bth}$ is well outside the perturbation distribution for soft quantization, but within the distribution for 4-bit HEQ: this result suggests that if we pick a random direction in weight-space and perturb a pretrained model along this direction, we reliably expect this direction to incur a greater loss than the one chosen by soft quantization, but not necessarily greater than the direction chosen by HEQ. Additionally, the distribution of $\frac{\Delta\Ltask}{\Delta\bth}$ for random perturbations has a lower average in the case of soft quantization than for HEQ. We infer that the displacement of weights is more aligned with flatter directions of the loss landscape for soft quantization than for HEQ, not only at the level of individual weight displacements, but also at the level of layerwise displacements. 

\noindent\textit{Conclusions and Future Directions.---}
We have introduced soft quantization, a method for compressing neural networks during fine-tuning via attractive couplings between weights. On ResNet-20/CIFAR-10, this method produces mixed-precision quantized models that outperform HEQ. The method relies on only two hyperparameters and is computationally efficient.

We have shown that soft quantization bridges the gap between compression and learning in a manner that is distinct from existing quantization paradigms, and as such it deserves broader investigation. An important direction is to assess its performance at larger scale and on more complex tasks, such as language models \cite{jin2024comprehensiveevaluationquantizationstrategies}, where comparable gains would give insights into the scale-dependence of redundancy exploitation via soft constraints. 
Another natural extension, which is of interest to researchers investigating the intersection of compression with phenomena such as transfer learning and continual learning, would be to vary the relationship between the pretraining and soft quantization datasets \cite{DBLP:journals/corr/abs-1805-08974,DBLP:journals/corr/HuhAE16}.
% Another natural extension is to vary the the relationship between pretraining and quantization datasets, for example bu deriving a compressed ``specialist" network from a more general pretrained network might experiment with using a specific subset of the training data during soft quantization. 
By explicitly linking compression to loss-landscape geometry, soft quantization provides a framework for probing the relationship between neural network compressibility and learning.
% Acknowledgments
\begin{acknowledgments}
We thank our colleagues W. Bialek, C. Goddard, F. Mignacco, V. Ngampruetikorn, L. Smith, and W. Zhong for helpful discussions and the National Science Foundation for support through the Center for the Physics of Biological Function (PHY--1734030). DJS was partially supported by a Simons Foundation Fellowship in the MMLS.

\end{acknowledgments}
% References - use BibTeX for best results
% Create a separate .bib file with your references
% \printbibliography
\bibliography{bibliography}
\newpage
\newpage
\FloatBarrier{}

\section{End Matter}
\begin{figure}[h!]
    \centering
    \includegraphics[width=\linewidth]{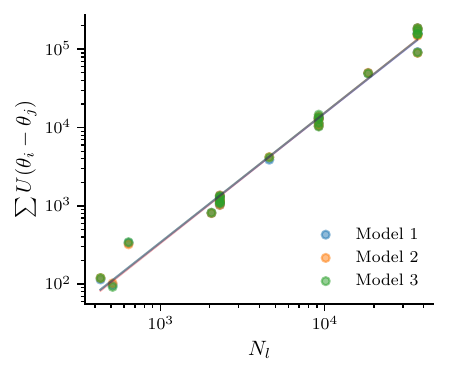}
    \caption{Power law between layer size and associated extensive ($h_l=1$) per-layer compression potential for the ansatz $w_l\sim\sigma_l$ for three distinct pretrained models. The exponent of this power law is approximately $1.66$. Average mean-squared error in this power law is 0.065.}
    \label{fig:powerlaw}
\end{figure}
\noindent\textit{Scaling of Potential.---}
If we require that the layer-wise compression potential scales with the number of layer parameters:
\begin{equation}
\frac{h_l \sum_{i \neq j} U_{w_l}\left(\theta_i^{(l)} - \theta_j^{(l)}\right)}{N_l} \approx \frac{h_l (w_lN_l^2)}{N_l}  \sim \text{constant}
\end{equation}
we infer that
\begin{equation}
h_l \sim \frac{1}{w_l N_l}.
\end{equation}
We assume that $w_l \sim\sigma_l$, which does not scale directly with $N_l$, so we are interested in a relationship between $h_l$ and $N_l$ of the form $h_l \sim N_l^{-\alpha}$. As shown in Figure \ref{fig:powerlaw}, at pretrained initialization we observe a power law between the extensive compression potential \begin{equation}
\sum_{i \neq j} U_{w_l}\left(\theta_i^{(l)} - \theta_j^{(l)}\right)
\end{equation} and layer size, and this power law suggests that $\alpha\approx0.66$.

\noindent\textit{Coefficients of Variation.---} Here we show the coefficients of variation across simulations for accuracy degradation and average bit-width. We note that these values are low when soft quantization outperforms 4-bit HEQ, indicating that performance and bit-width are relatively consistent for a given $h$ and $w$.
\begin{figure}[htbp]
    \centering
\includegraphics[width=\linewidth]{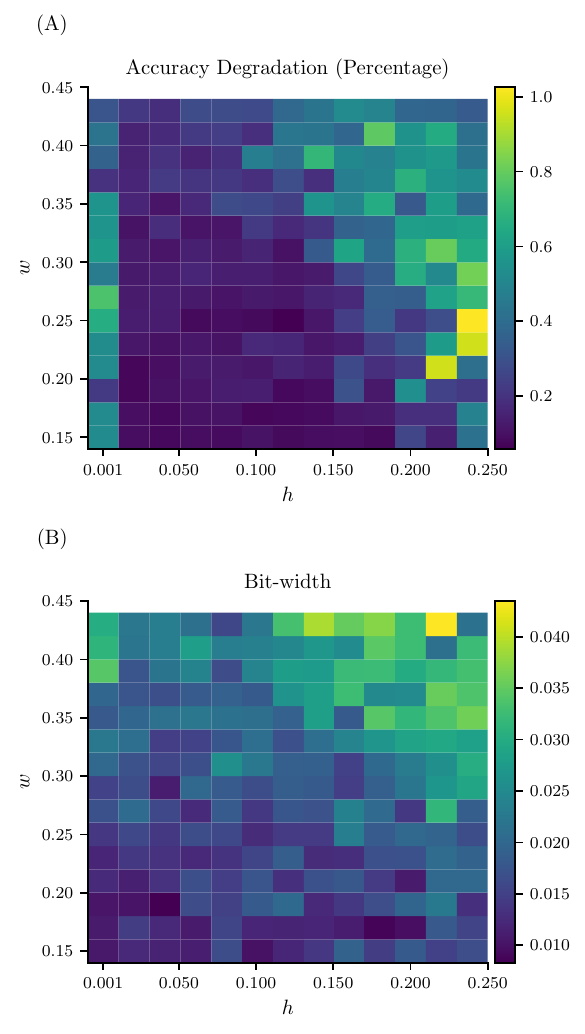}
\caption{Coefficients of variation for (A) test accuracy degradation and (B) average bit-width.}\label{fig:coeff_var}
\end{figure}
\newpage
\noindent\textit{Soft Quantization Before Model Refinement.---}
Here we present the same data as shown in Figure \ref{fig:Main_Scatterplot}(A)--(B), but before the optional refinement step (reassignment of small clusters, see Table~\ref{tab:soft_quant_algorithm}).
\begin{figure}[!htbp]
    \centering  \includegraphics[width=\linewidth]{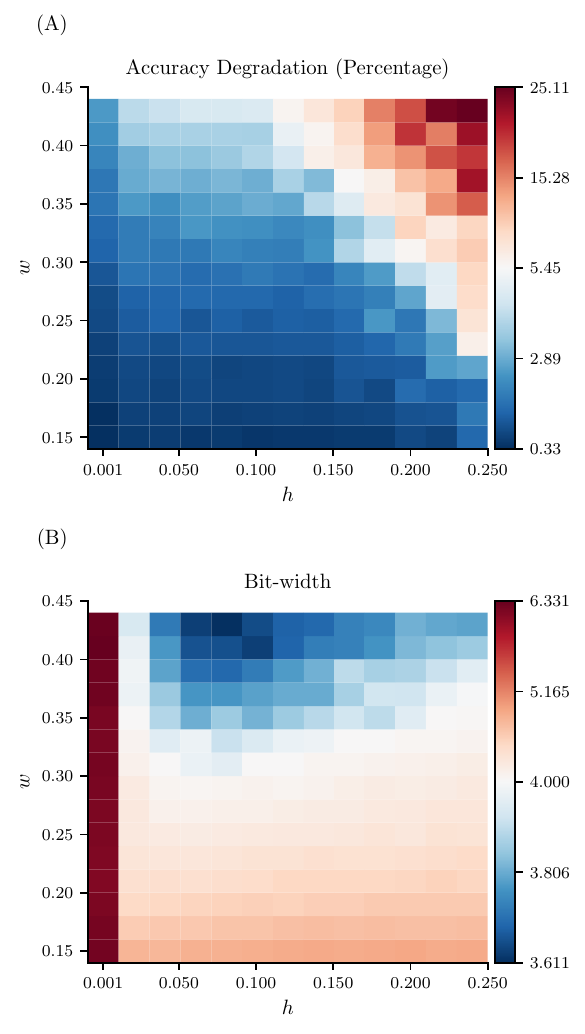}
\caption{Soft quantization statistics for test accuracy degradation (A) and average bit-width (B) before refinement. For small $h$ ($h=0.001$), the effect of the potential is minor, and the upper-bound of 6.331 on the average bit-width is an artifact of how we identify unique clusters after re-training, i.e.\ by binning weights at 7-bit precision. For hyperparameter settings with lower bit-widths (in particular the settings outperforming HEQ), this identification process is a formality which does not affect model accuracy, as 7-bit precision is well above the resultant bit-width.}
    \label{fig:clamped_sq}
\end{figure}
\end{document}